# Title: The Scope and Limits of Simulation in Cognitive Models

## Authors:[1]

Ernest Davis, Dept. of Computer Science, New York University, New York, NY, davise@cs.nyu.edu, http://www.cs.nyu.edu/faculty/davise

Gary Marcus, Dept. of Psychology, New York University, New York, NY, gary.marcus@nyu.edu, http://www.psych.nyu.edu/gary/

## December 10, 2014

## Abstract:

It has been proposed that human physical reasoning consists largely of running "physics engines in the head" in which the future trajectory of the physical system under consideration is computed precisely using accurate scientific theories. In such models, uncertainty and incomplete knowledge is dealt with by sampling probabilistically over the space of possible trajectories ("Monte Carlo simulation"). We argue that such simulation-based models are too weak, in that there are many important aspects of human physical reasoning that cannot be carried out this way, or can only be carried out very inefficiently; and too strong, in that humans make large systematic errors that the models cannot account for. We conclude that simulation-based reasoning makes up at most a small part of a larger system that encompasses a wide range of additional cognitive processes.

## 1. Introduction

In computer science, virtually all physical reasoning is carried out using a physics engine of one kind or another. Programmers have created extremely detailed simulations of the interactions of 200,000,000 deformable red blood cells in plasma (Rahimian, et al., 2010); the air flow around the blades of a helicopter (Murman, Chan, Aftosmis, & Meakin, 2003); the interaction of colliding galaxies (Benger, 2008); and the injuries caused by the explosion of an IED under a tank (Tabiei & Nilakantan, undated). Software, such as NVidia PhysX, that can simulate the interactions of a range of materials, including rigid solid objects, cloth, and liquids, in real time, is available for the use of game designers as off-the-shelf freeware (Kaufmann & Meyer, 2008). In artificial intelligence (AI) programs, simulation has been used for physical reasoning (Johnston & Williams, 2007), (Nyga & Beetz, 2012), robotics (Mombauri & Berns, 2013), motion tracking (Vondrak, Sigal, & Jenkins, 2008), and planning (Zicker & Veloso, 2009).

So it is perhaps hardly surprising that some have come to contemplate the notion that human physical reasoning, too, might proceed by physics engine. Battaglia, Hamrick, and Tenenbaum (2013, p 18327I, for example, recently suggested that human physical reasoning might be based on an "intuitive physics engine," a cognitive mechanism similar to computer engines that simulate rich physics in video games and graphics, but that uses approximate, probabilistic simulations to make robust and fast inferences.

---

[1] The order of authors is alphabetical.

They went on to conjecture that most intuitive physical reasoning is carried out using probabilistic simulation:

Probabilistic approximate simulation thus offers a powerful quantitative model of how people understand the everyday physical world. This proposal is broadly consistent with other recent proposals that intuitive physical judgments can be viewed as a form of probabilistic inference over the principles of Newtonian mechanics.

Similarly, Sanborn, Masinghka, and Griffiths (2013) recently proposed the strong view that "people's judgments [about physical events such as colliding objects] are based on optimal statistical inference over a Newtonian physical model that incorporates sensory noise and intrinsic uncertainty about the physical properties of the objects being viewed … Combining Newtonian physics with Bayesian inference, explaining apparent deviations from precise physical law by the uncertainty in inherently ambiguous sensory data, thus seems a particularly apt way to explore the foundations of people's physical intuitions."

In this paper, we consider this view, as well as a weaker view, in which simulation as viewed a key but not unique component in physical reason. Hegarty (2004) for example argued that, "Mental simulations ... can be used in conjunction with non-imagery processes such as task decomposition and rule-based reasoning " Along somewhat similar lines, Smith et al. (2013) argued that physical reasoning is generally carried out using simulation, but admit the possibility of exceptions:

In some specific scenarios participants' behavior is not fit well by the simulation based model in a manner suggesting that in certain cases people may be using qualitative, rather than simulation-based, physical reasoning.

Our purpose in this paper is to discuss the scope and limits of simulation as a cognitive model of physical reasoning. To preview, we will we accept that simulation sometimes plays a role in some aspects of physical reasoning, but also suggest that there are some severe limits on its potential scope as an explanation of human physical reasoning. We will suggest that in many forms of physical reasoning, simulation is either entirely inadequate or hopelessly inefficient.

1. Profound systematic errors are ubiquitous in physical reasoning of all kinds. There is no reason to believe that all or most of these can be explained in terms of physics engines in the head.
2. Subjects' accuracy in carrying out tasks varies very widely with the task, even when the same underlying physics is involved. This diminishes the predictive value of the theory that subject use a correct physical theory.
3. In some forms of physical reasoning, simulation-based theories would require that naïve subjects enjoy a level of tacit scientific knowledge of a power, richness, and sophistication that is altogether implausible.

## 2. Simulation-based physical reasoning

Roughly speaking, there have been two distinct classes of work in the area of simulation and human physical reasoning.

## 2.1 Depictive models

One strand, published mainly over the 1990s and prior to 2005, couched important aspects of physical reasoning in terms of visualizations of behavior evolving over time. Figure 1 illustrates three typical example from the earlier literature:.

Hegarty (1992) studied how people infer the kinematics of simple pulley systems (3 pulleys, 1 or 2 ropes, 1 weight) from diagrams showing a starting state. Her primary data was eye fixation studies (i.e. the sequence in which subjects look at different parts of the diagram or the instructions), though accuracy and reaction times were also reported. Hegarty proposes a weak theory of simulation, in which subjects simulation each pairwise interaction between components of the system and then trace a chain of causality across the system, rather than attempting to visualize the workings of the system as a whole.

In Schwartz and Black (1996) subjects solved problems involving the motion of two gears. Two adjacent gears of different sizes are shown with either a line marked on both or a knob on one and a matching groove on the other. The subjects were asked whether, if the gears were rotated in a specified direction, the lines would come to match, or the knob would meet the groove. In all experiments, both accuracy and latency were measured, but most of the conclusions were based on latency. The primary purpose of the set of experiments as a whole was to compare the use of two possible reasoning strategies; on the one hand, visualizing the two gears, rotating synchronously; on the other hand, comparing the arc length on the rim of the two gears between the contact point in the starting state and the lines/knob/groove. (The interlocking of the gears enforces the condition that equal arc lengths of the rim go past the contact point; hence, the two markings will meet just if the arc lengths are equal.) It was conjectured that the first strategy would require a cognitive processing time that increased linearly with the required angle of rotation, but that the second strategy would be largely invariant of the angle of rotation; and the experimental data largely supported that conjecture. Various manipulations were attempted to try to get the subjects to use one strategy or another. In one experiment they were specifically instructed to use a particular strategy. In another, they were presented alternately with a realistic drawing or with a schematic resembling a geometry problem; the former encouraged the use of visualization, the latter encouraged the comparison of arc length.

Similarly, Schwartz (1999) studied the behavior of subjects who are trying to solve the following problem. "Suppose there are two glasses of the same height, one narrow and one wide, which are filled with water to equal heights. Which glass has to be tilted to a greater angle to make it pour?" Schwartz reports that subjects who answer the question without visualizing the two glasses almost always get the answer wrong (19 out of 20 subjects). However, if they visualize tilting the glasses, they are much more successful. Schwartz further tried a number of manipulations to determine when the subjects use a kinematic model and when they use a dynamic model; frankly, the relation between this distinction and the structure of his experiments is not always clear to us. The data that he used was the subjects' answers.

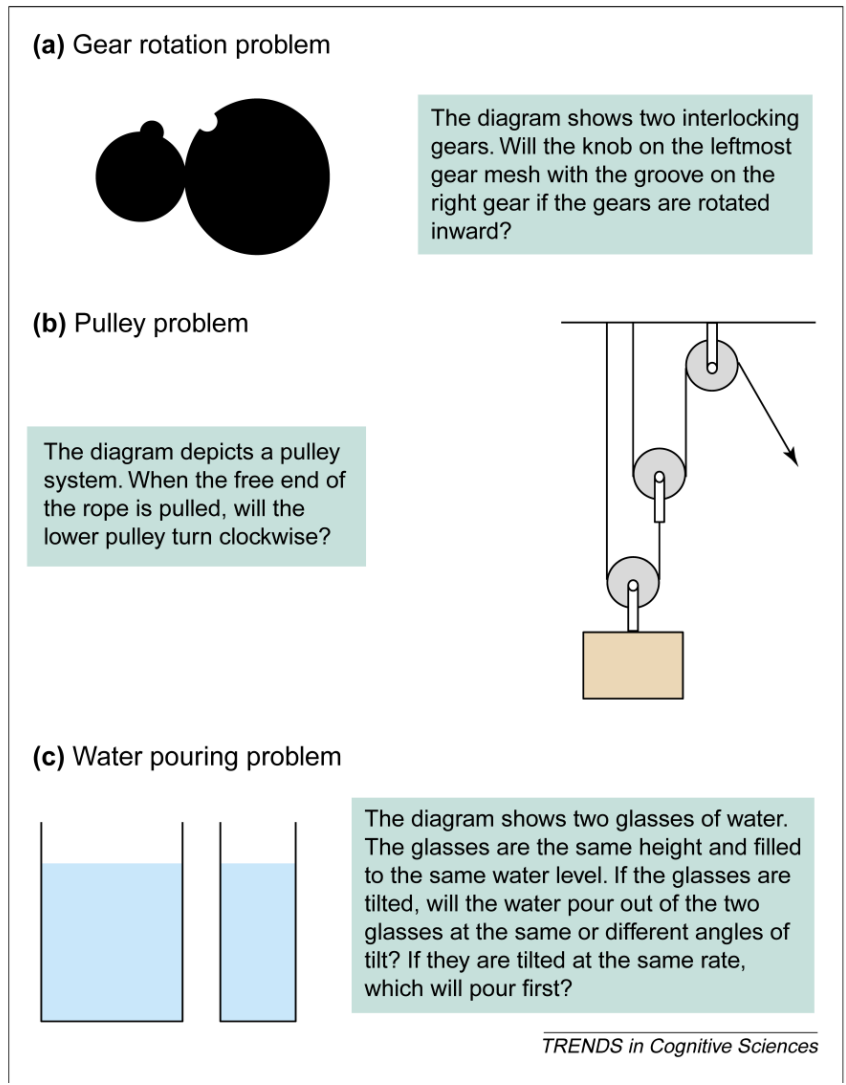

Figure 1: Experiments. From (Hegarty, 2004)

## 2.2 Newtonian physics engines

More recent work has been couched in terms of what one might describe as a "physics engine in the head" approach (phrase due to Peter Battaglia, pers. comm.) A "physics engine" is a computational process analogous to the physics engines used in scientific computations, computer graphics, and computer games. Broadly speaking, the physical theory that is incorporated in the engine is expressed in the form of update rule, that allows the engine to compute the complete state of the world at time T+Δ given a complete specification of its state at time T, where Δ is a small time increment. (We will discuss the significance of "complete" below.) In any particular situation, the input to the engine is the complete state of the world at time T=0; it then uses the update rule to compute the state of the world at time Δ from its state at time 0; to compute the state of the world at time 2Δ from its state at time Δ and so on. Thus, it computes an entire trajectory.

Many variants of this general idea are possible. There may be exogenous events that occur over time, such as the actions of an player in a game; in that case, the update function may have to take these into account. It may be possible to extrapolate the trajectory of the world to the next "interesting" event rather than using a fixed time increment Δ; for instance, in the bouncing ball experiment of (Smith, Dechter, Tenenbaum, & Vul, 2013) described below, the engine might well goes from one bounce to the next without calculating intermediate states, except to check whether the path crosses one of the target regions. Battagila, Hamrick, and Tenenbaum (2013) suggest that the internal engine is "approximate: In its mechanics rules and representations of objects, forces, and probabilities, it trades precision and veridicality for speed, generality, and the ability to make predictions that are good enough for the purposes of everyday abilities". However the inference engine that they actually use in their model is not approximate in this sense.

Most important, if either the starting state or the update rule is partially specified but can be viewed as following a probabilistic distribution, then one can generate a random trajectory corresponding to that distribution by random sampling; this is *probabilistic* or *Monte Carlo* simulation. This partial specification of the input situation is usually attributed to limits on the precision of perception or to partial information of some other kind about the starting situation. Probabilistic models of this kind are known as "noisy Newton" models, since they follows Newtonian mechanics (or whatever exact scientific theory applies) with the addition of some random noise.

In this more recent line of work, subjects typically view and interact with a computer simulation rendered on a two-dimensional monitor. In some experiments (see Figure 2), the monitor displays an two-dimensional rendering of a three-dimensional situation; in the other experiments, the simulation in question is itself two dimensional . (In the depictive work in earlier decades, participants were generally shown static pictures. Experiments in physical reasoning where subjects interact with or view live, 3-dimensional physical situations seem to be mostly restricted to studies with children.)

To take one example (left panel of Figure 2), Battaglia, Hamrick, and Tenenbaum (2013) carried out one experiment in which participants were shown a tower of blocks. Participants were asked to predict whether a tower was stable and, if not, in which direction it would fall. In other experiments (right panel of figure 2) subjects were shown a table with a number of towers of red and green blocks in various positions. Participants were told that the table would be struck at a specified point, and they were asked whether more red or more green blocks would fall off. Responses were consistent with a "noisy Newton model" in which a subject applies the Newtonian theory of rigid solid objects, and carries out probabilistic simulation, where the probabilistic element corresponds to uncertainty in the positions of the blocks

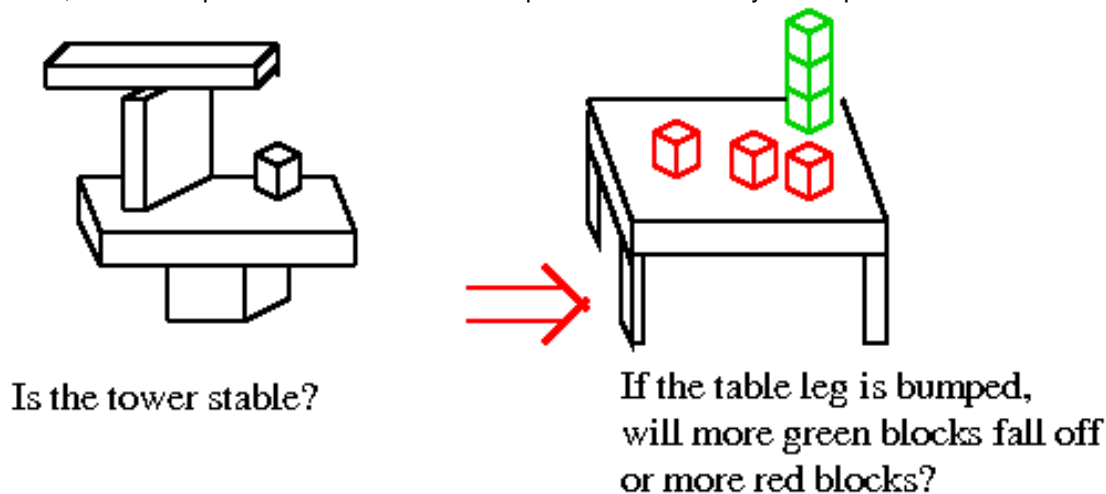

**Figure 2: Experiments in (Battaglia, Hamrick, and Tenenbaum, 2012)**

In another study in this strand, Smith and Vul (2013) carried out an experiment in which participants catch a bouncing ball with a paddle. The trajectory of the ball is initially visible; then one side of the screen becomes occluded, and the subjects must move the paddle to a position where it will catch the ball after it has bounced around (figure 3). The data analyzed was relation of the subjects' placement of the paddle to the actual trajectory of the ball. They were able to match the data to a "noisy Newton" model in which both the perception of the ball's position and velocity and the calculation of the result of a bounce were somewhat noisy. They additionally posited a "center bias", with no theoretical justification, which they attributed to the subjects' prior expectations about the position of the ball.

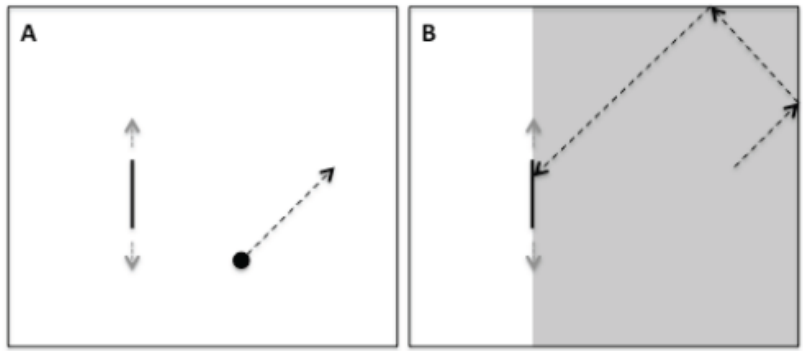

**Figure 3: Diagram of a trial. (A) The ball moves unoccluded in a straight line. (B) Once the field is occluded, the ball continues to move until caught or it passes the paddle plane. From (Smith and Vul, 2013)**

Experiments in Smith et al. (2013) likewise tested people's ability to predict the trajectory of a bouncing on a table with obstacles A green and a red target are marked on the table, and subjects are asked which of the two targets the ball will reach first. As they watched the simulated ball bounce, subjects were able to continuously make their best guess as to the answer, and to change their prediction when necessary. The data considered was thus the time sequence of guesses. They found that in most cases subjects' answers fit well to a noisy Newton model similar to that of (Smith & Vul, 2013) (they additionally posited a rather complex decision model of how subjects chose their answer.) The exceptions were cases where the correct prediction could be made purely on the basis of topological constraints; i.e. cases where any possible motion would necessarily reach one region before the other. In those cases, subjects were able to find the correct answer much earlier than their model predicted (figure 4).

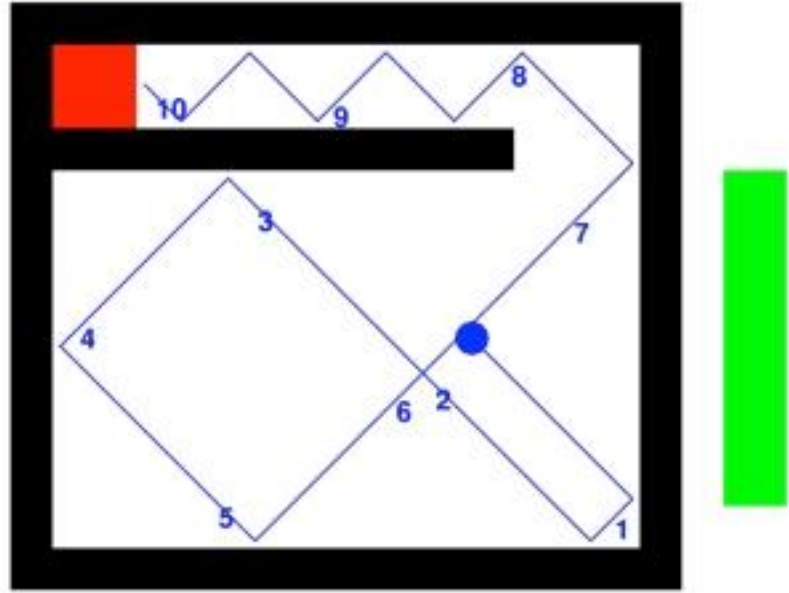

**Figure 4: Simulating a bouncing ball: An example that can be solved using qualitative reasoning. From (Smith, Dechter, Tenenbaum, and Vul, 2013)**

Smith, Battaglia, and Vul (2013) carried out experiments testing how well subjects could predict the trajectory of a pendulum bob if the pendulum is cut while swinging. Replicating the negative results of (Caramazza, McCloskey, & Green, 1981), they found that when subjects were asked to draw the expected trajectory, they did extremely poorly, often generating pictures that were not even qualitatively correct. However, if subjects were asked to place a bucket to catch the bob or to place a blade, they were much more accurate. (Figure 5)

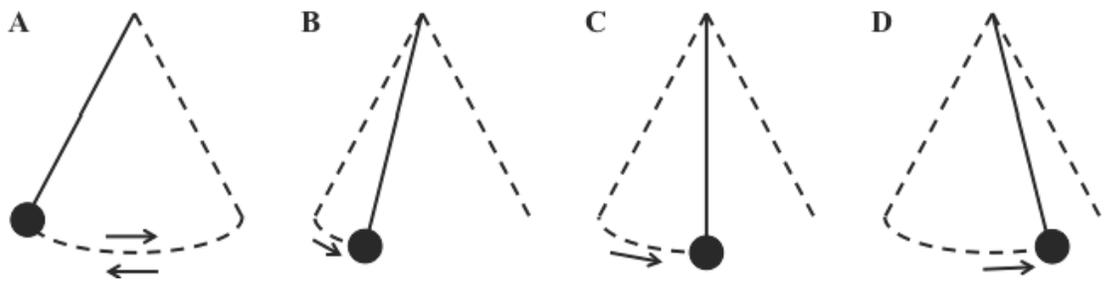

**Figure 5: The four pendulums in the diagram task. Participants were asked to draw the expected path of the ball if the pendulum string were cut at each of the four points. (Figure 2 of (Smith, Battaglia, and Vul, 2013))**

(Sanborn, Mansinghka, & Griffiths, 2013) , (Gerstenberg et al., 2014), (Sanborn, 2014), (Gerstenberg & Goodman, 2012), and (Smith & Vul, 2014) study human performance on a variety of tasks involving colliding balls, including prediction, retrodiction, judgment of comparative mass, causality, and counterfactuals. In all cases, the data used are the subjects' responses and in all cases they found a good fit to a noisy Newton model.

Two further studies also fall broadly in the category of "noisy Newton", though in one way and another they are not actually based on realistic physics. In the experiments described in (Teglas, et al., 2011) twelve-month old babies were shown a display of objects of different shapes and colors bouncing around inside a container with holes. The inside of the container was then occluded, and, after a delay, they could see that an object escaped the container. By varying the number of objects of each type, the initial distance from the objects to the holes, and the time delay before the escape, the experimenters were able to create scenarios with varying levels of probability, as measured in their model. As usual, babies' judgment of the likelihood of the scenario they have seen was measured in terms of the staring time (longer staring times correspond to less likely outcomes); the data in fact showed an astonishingly precise linear negative relation between the probability in the model and the babies' staring time.[2] The physical model posited in this paper is that the objects inside the container move entirely randomly (Brownian motion), subject to the constraint that they can only exit the container through its holes. This physical theory is thus rather different in flavor from the other physics-engine based theories considered earlier. Here, probabilistic variation is a central feature of the physical theory, rather than a result of limited perception or a minor modification of a dynamic theory, as in the other works in this school

Ullman et al. (2014) study how subjects can learn the physical laws governing a new domain (a collection of different kinds of "hockey pucks" moving and interacting on a horizontal surface). They propose a complex, multi-level Bayesian model. Since this is primarily a study of learning a new domain, rather than using pre-existing knowledge, it is not entirely comparable to the other studies discussed in this section.

## 2.3 Progress?

Before getting to the central thrust of our critique, it is worth noting that in some ways, the newer literature is a step back relative to the earlier literature. Although the newer work has the advantage of being much more mathematically precise, it is far less explicit in terms of actual computational mechanism. The earlier work delved directly into psychological mechanisms, using a wide range of data sources, such as latency data (Schwartz and Black, 1996), eye fixation data (Hegarty & Sims, 1994), cross-subject correlation with other measures of spatial reasoning ability (Hegarty & Sims, 1994), measures of task interference (Sims & Hegarty, 1997); and protocol analysis (Schwartz & Black, 1996). In contrast, the newer, physics-engine inspired work, has relied almost exclusively on accuracy data, with the risk that many different mechanisms might actually yield the same data. To take one example, the problems in (Smith, Battaglia, & Vul, 2013) of cutting a pendulum involve reasoning about the pendulum motion and reasoning about the free fall motion. These could be computed using a physics engine or they could be computed using an analytic expression. The angle of the bob while swinging as a pendulum is approximately a trigonometric function of time, or more precisely an elliptic function of time; and the free fall motion is a parabolic motion function of time. It might conceivably be possible to distinguish the use of a physics engine from the use of an analytic solution using data such as latency; but they cannot be distinguished in terms of subject responses, since the two approaches give the same results.

---

[2] Even if the model were an exactly precise account of the babies' judgment of likelihood, there is no theoretical reason to expect a linear relation between the babies' judgment of likelihood and their staring times. All that the theory predicts is that the staring time will be a decreasing function of the judged likelihood.

Furthermore, in the newer work there has been little commitment to specific psychological processes Indeed, in at least some of the recent work (e.g. (Sanborn, 2014, p. 1)), the models are put forward only as models at the "computational level" (Marr, 1982), with no suggestions that they describe the actual reasoning process.

As a result some of the process results found by the old works raise difficulties for the newer work For example, Hegarty (1992) found evidence that subjects do not visualize a multi-part process as a whole; instead, her suggestion was that human reasoners trace causal chains, in a piecemeal fashion. However, such a strategy can hardly be used in the Newtonian model of falling towers of blocks proposed in (Battaglia, Hamrick, & Tenenbaum, 2013). For problems like these, a physics engine based on Newtonian must necessarily calculate the trajectories of all the blocks in parallel, because the force exerted at one end of a tower can affect the force exerted at another, and, as the tower collapses, the forces change continuously over time. It is not clear that Battaglia's account can be reconciled with Hegarty's earlier results. More broadly, the strict focus on models at Marr's computational level may led to many such conflicts, especially when considerations such as working memory and attention are factored in.

The work in the recent tradition, at least thus far, has been more narrow in scope, limited to situations with rigid solid objects, and often quite specialized idealizations of these, such as perfectly elastic, spinless, collisions of balls in the horizontal plane. Even (Battaglia, Hamrick, & Tenenbaum, 2013), which has the least specific assumptions of the work in this domain, limits itself to towers of rectangular blocks that are initially aligned with the coordinate axes; and considered perceptual imprecision in the exact placement of the object but not in its orientation or shape. The problems considered in the earlier work typically had a different flavor: more varying physical phenomena (strings, pulleys, gears, liquids) but very few degrees of freedom and numerical parameters . For example in Schwartz and Black's gear rotation problem, motion occurs along a single degree of freedom (the coupled rotations of the two gears) and the problem specification involves three significant numerical parameters (the ratio of the diameters of the gears and the initial angles of the knob and the notch.) By contrast, each of the tower problems in (Battaglia, Hamrick, & Tenenbaum, 2013) involves motion with 60 degrees of freedom. The literature on errors in human physical reasoning, discussed below in section 4.1, deals with a much broader range of physical phenomena than either of these.

One important general conclusion that emerges from both lines of work, is that there is probably no "one size fits all" model of human physical reasoning; rather, it appears likely people have a number of different strategies they can use, depending on the circumstances

## 3. Inherent limitations of simulation

However one approaches the simulation hypothesis, there is a deeper issue: simulation simply isn't as general a technique as it may initially seem, especially to a non-expert. It is easy for a non-professional to overestimate the state of the art of physical simulation, and assume that there is a plug-and-play physics engine that works for pretty much any physical situations. If computer scientists actually knew how to build a comprehensive physics engine, the idea that humans might have an instantiation of such a thing might seem more plausible. But the reality is that physics engines of the sort that exist in 2014 are brittle things, when it comes to anticipating the full complexities of the real-world. Plug-and-play engines capture only narrowly-defined environments; more sophisticated applications require hard work from experts. A few seconds of realistic CGI in a disaster film may well require several person-days of work; an accurate and complex scientific computation may require several person-months. As simulation expert Nils Thuerey (personal communication) put it, "There are ... inherent difficulties with these simulations: we are still very far from being able to accurately simulate the complexity of nature around us. Additionally, the numerical methods that are commonly used are notoriously difficult to fine-tune and control." Plug-and-play physics engines are also subject to bugs and anomalies,[3] and may require careful, post hoc parameter setting to work correctly. In a systematic evaluation of seven physics engines, Boeing and Bräunl (2007) found that all seven gave significantly and obviously erroneous answers on certain simple problems involving solid objects.

Even within automated reasoning of the sort found in artificial intelligence, there at least a dozen serious challenges to using simulation as a full-scale solution to physical reasoning, as we discuss in a forthcoming review (Davis & Marcus, to appear). And many of the challenges that we discuss there lead to analogous difficulties for any pure-simulation account of *human* physical reasoning as well. We discuss the six most important of these challenges.

## 3.1 Finding an appropriate modeling approach

For a programmer building a simulation, the first hurdle in implementing a simulator is developing a domain model. In some cases, such as pendulums and blocks, this is well understood. However finding an appropriate model can often be difficult, even for familiar objects, materials, and physical processes. Consider, for instance, cutting materials with tools. An ordinary household has perhaps a dozen kinds of tools for cutting: a few kinds of kitchen knives; more specialized kitchen equipment such as graters and peelers; a few kinds of scissors; a drill, a saw, a lawn mower, and so on. (A specialist, such as a carpenter or a surgeon, has many more.) Most people understand how they should be used and what would happen if you used the wrong tool for the material; if, for example, you tried to cut firewood with a scissors. But it would be hard to find good models for these in the physics or engineering literature. Where the best minds in computer science and engineering haven't yet constructed detailed simulations of these sorts of situations, it might be unrealistic to presume that brains are solving routinely solving a vast array of such problems via simulation alone.

## 3.2 Choosing an idealization

Virtually all actually-implemented simulations represent idealizations; in some, friction is ignored, in others three dimensions are abstracted as two. In most situations, many different idealizations are possible; and the idealization should be chosen so that, on the one hand, the calculation is not unnecessarily difficult, and on the other, the important features of the situation are preserved. Consider, for instance, the simulation of a pendulum on a string. As we have discussed Smith, Battaglia, and Vul (2013) generated simulations in which the bob first swings on the string and then, after the string is cut, flies freely through the air. In all likelihood, the bob was modeled throughout as a point object, moving under the influence of gravity, and the string was modeled as a constraint that requires the bob to move on a circular path. The cutting of the string was modeled as an instantaneous elimination of this constraint, with the assumption that the instantaneous velocity of the bob is unchanged.

Other scenarios for a bob on a string are more complex. A bob may swing in a horizontal circle; spin on the axis of the string; rotate about its center of mass like a yo-yo, or fly through the air (figure 6). The string itself may be taut, loose, tangled, knotted, or twisted; it may get in the way of the bob; it may even unravel or snap. Although these behaviors are familiar to anyone who has spent time playing with objects on strings, few if any existing physics engines support any but the taut and loose conditions of the string and perhaps snapping.

---

[3] Many can be found on YouTube by searching on the name of the engine plus "Bug"; e.g. https://www.youtube.com/watch?v=Jl-A5YaiWi8 shows a bug with gyroscopic motion in ODE and https://www.youtube.com/watch?v=vfl33Tn0pYc shows a number of bugs in Skate 3.

Efficient reasoning about these different possible behaviors of the string and the bob requires using a variety of different idealizations. Space can be two-dimensional or three-dimensional. A bob can be idealized as a point object, a rigid object, or an elastic object. A string of length L can be idealized as an abstract constraint restricting the motion of the bob; a one-dimensional curve of length L, with or without mass; or a three-dimensional flexible object, either uniform or with some internal structure (e.g. twisted out of threads or a chain of links.)  Cutting the string may be instantaneous and involve no dissipation of energy; in a more realistic model, it will require finite time and involve a small dissipation of energy. Influence on the system can be limited to gravity, or can include friction and air resistance. In looking at any one simulation that has been well-worked out, it is easy to lose sight of how much careful work goes on in choosing the right idealization; as yet there is no algorithmic way to guarantee an efficient solution for arbitrary problems. Using the most realistic model possible is no panacea; highly realistic models both require more laborious calculation and more detailed information.

Useful idealizations can, indeed, be very far from the underlying physical reality. As we have discussed, Teglas et al. (2011) model balls bouncing inside a container in terms of Brownian motion; i.e. each ball moves on a random path, uninfluenced by the other balls. This is indeed a common idealization in statistical mechanics (where it is much more nearly correct, due to the enormous number of particles – but not at all a reasonable model of the dynamics of an individual ball. The motions of an actual ball in this situation are due to collisions with other balls, not due to an autonomous random process.

 A simulation-based model of human reasoning therefore cannot rely on simply running a uniquely-defined model for each type of object or substance. Rather, objects come with a collection of models; and choosing the right model for each object in a given situation is itself a complex reasoning task.

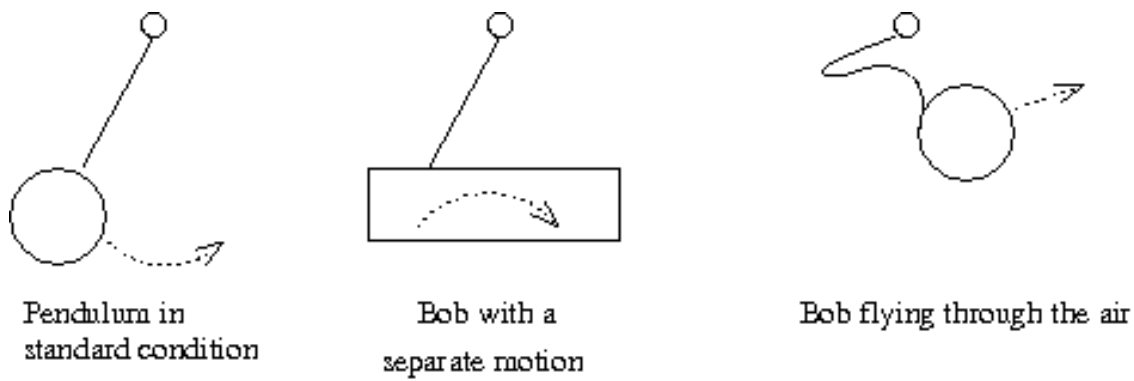

Figure 6: A pendulum in various conditions

## 3.3 Rapid, approximate inferences

The output of a computer simulation is invariably precise, bit not always accurate.  Human physical reasoning seems to have a different character, speedy, yet less precise.

Correspondingly, human reasoners often seem to have little need for the level of precision that a simulation provides. If you ride a bicycle on a bumpy road while carrying a half-full closed water canteen, all that matters is that the water stays inside the canteen, not the trajectory of the water splashing inside the canteen. Humans often seem to be able short cut these situations, drawing rapid inferences.

Although the exact workings of human physical reasoning remain unknown, there are many kinds of rules that in principle could allow quick inference or quick transference of results from a known situation to a new one. *Invariance under time and space*: If a mechanism worked in a particular way at home on Monday, it will work in the same way at work on Tuesday. *Invariance under irrelevant changes*: If a jar fits on a shelf, and you fill it with pennies, it will still fit on the shelf. *Invariance under changes of scale* (for certain physical theories, such as kinematics): A large pair of gears works in the same way as a much smaller scale model. *Approximation*: If a jug holds a gallon of water, then another jug of similar dimensions and shape will hold about a gallon. *Ordering on a relevant dimension*: If a toy fits in a box, then it will fit in a larger box, under a proper definition of "larger" (Davis, 2013). *Decomposition:* If a system consists of two uncoupled subsystems, then one can reason about each separately.

To take another example: suppose a subject is shown a picture of one of Hegarty's pulley problem, as in figure 1 above, and asked whether, if the weight is made heavier, more or less force would be required to lift it through the pulley system. One possible approach would be to calculate the force exactly for a number of different weights and compare them. But a simpler approach, and we conjecture the approach that most people would use, is to use a general rule:, in any mechanical system for lifting an object, the force that must be applied is proportional to the weight to be lifted.

There can also be *rules of thumb*: useful if imperfect generalizations for common cases. For instance, if you spill a cup of coffee in your office, it should not be necessary to resort to simulation to infer that the coffee will not end up in some other office. Rather, one can use a rule of thumb that a small amount of liquid dropped from a moderate "height onto a horizontal surface will end up not very far from the point where it was dropped, where "small amount", "moderate height", and "not very far" have some very approximate quantitative measurements.

These alternative forms of inference may do well under circumstances where simulation scales badly. Consider the following scenario: you put a number of objects in a box, close the lid, and shake it up and down violently. We now wish to infer that the objects are still in the box. Simulating the motion of the objects will become rapidly more complicated as the number of objects increases, the shapes of the objects become more complex, and the shaking of the box becomes more complex and violent. By contrast a single simple rule, "An object in a closed container remains in the container" suffices to carry out the inference. Our best guess is that such rules are an important part of the fabric of human physical reasoning.

## 3.4 Extra-physical information

In some cases, reasoning about a physical system can be carried out more reliably and effectively using extra-physical information. Suppose that you see a baseball pitcher throw a ball. If you simulate the motion of the ball, using the best information you can gather about the angle, velocity, and so on, of the ball when it left his hand, and factor in the imprecision of this information, you will predict that it has a rather low probability of ending up anywhere close to the batter. You would obtain better results — predicting that the ball will end up close to the strike zone, just inside it or just outside — by instead relying on the known accuracy of the pitcher, plus quite specific information about the state of play and the pitcher's willingness to risk an out-of-strike zone pitch rather than a hit. Pure physical simulation, constrained by real-world measurement error, might produce far less accurate results.

## 3.5 Incomplete information

Carrying out a physical simulation is generally only possible if the geometric and physical characteristics of the initial condition are known precisely. In many common real-world situations, humans often perform physical reasoning on the basis of partial, sometimes extremely limited, information.

Perception, for example, may be imperfect or incomplete. For instance, an object may be partially occluded. (An opaque object always self-occludes its own far side from the viewer.) Knowledge of the physical situation may come from natural language text or sketches. Knowledge of aspects of the situation may come from inference; for example, if you see someone attempt unsuccessfully to pick up a suitcase, you can infer that the suitcase is unusually heavy; you can then use that inference for future prediction. Or the precise details may not have been determined yet. For instance, suppose that you are going to the furniture store to buy a dining room table. You can reason that you will not be able to carry it home on foot or riding a bicycle, even though you have not yet chosen a particular table.

Of course, no representation is truly complete or entirely precise; in any representation, some aspects are omitted, some are simplified, and some are approximated. However, the simulation algorithm requires that the initial conditions of the scenario be *fully specified relative to a given level of description*. That is, the representational framework specifies some number of critical relations between entities and properties of entities. A complete representation of a situation relative to that framework enumerates all the entities that are relevant to the situation, and specifies all the relations in the framework that hold between those entities. The description must be detailed and precise enough that the situation at the next time step is likewise fully specified, in the same sense

In many cases people are able to carry out physical reasoning on the basis of information that is *radically incomplete.* For example, suppose that you have an eel inside a closed fish tank and you wish to infer that it remains in the tank. If we are to solve this problem by probabilistic simulation, we would need, first to understand how an eel swims, and second to simulate all kinds of possible motions of the eel and confirm that they all end with the eel inside the tank. If we do not know the mechanisms of swimming in eels, then the only way to use probabilistic simulation is to generate some random distribution of mechanisms that eels might use. Clearly, this is not psychologically plausible.

## 3.6 Tasks other than prediction

Simulations are most readily used for the task of generating predictions, such as where a pendulum will be at a given instant. There are, however, many other important kinds of physical reasoning tasks, to which simulation-based techniques are in general less well suited.  These include: interpolation between two states at two different times; planning; inferring the shape of an object; inferring the physical properties of an object; design; and comparative analysis. All of these kinds of tasks arise constantly for people in ordinary situations; a cognitive theory of physical reasoning must therefore account for all of them. A system could only capture prediction would miss much of the richness of human physical reasoning.

# 4. Further limitations of simulation in cognitive modeling

Even where simulation could be used in principle, there is often reason to doubt that is used in practice.

## 4.1 Systematic errors in physical reasoning

> "How should I answer these questions — according to what you taught me, or how I usually think about these things?"
> — Harvard physics student, confronted with David Hestenes' test of basic physics concepts (Lambert, 2012)

There is an enormous literature demonstrating that both naïve and educated subjects make systematic, large errors in simple physical reasoning tasks. We have already discussed some of these. Tasks in which well-documented errors commonly occur include:

1. Predicting the trajectory of an object that has been moving in a circular path and is now released (McCloskey, 1983).
2. Drawing the predicted trajectory of bob on a pendulum that is cut (Smith & Vul, 2013)
3. Predicting the behavior of balance beams (Siegler, 1976).
4. Generating explanations for the workings of familiar mechanisms, such as bicycle gears (Keil, 2003).
5. Judging the relation between the masses of two colliding objects based on seeing a collision between them (Gilden & Proffitt, 1994).
6. Predicting whether two objects will in fact collide (Levillain & Bonatti, 2011)
7. Predicting whether another person can see themselves in a mirror (Lawson, 2012)
8. Predicting the behavior of a wheel (Profitt, Kaiser, & Whelan, 1990)

The list could be extended at very great length. These kinds of errors are inherently challenging for theories of simulation based on correct physical theories. In a few cases such as (5) above, it is possible to explain the errors in terms of a "noisy Newton" theory as the consequence of taking perceptual noise into account (Sanborn, 2014). In most other cases, no such explanation has been offered and in many no such explanation seems to be possible.

Although some of these might be individually explained away, as errors of misinterpretation, in our view such errors are simply too ubiquitous to be explained away. And although "Bayesian" approaches to cognition often presume optimality as a kind of default, there is no particular reason that human cognitive processes, developed by evolutionary processes, many over only a comparatively short time in evolutionary terms, *should* be error-free or otherwise optimal (Marcus, 2008)

Even sophisticated and experienced subjects, with time to think, can make elementary errors. Haroun and Hestenes (1985) found dismal level of performance in very simple problems involving force among college students, both on entrance to a freshman physics class, and after completing the course. Such errors on simple problems can be made, not only in a classroom setting with theoretical problems, but in a real-world setting, where the most serious possible consequences at stake, to a subject with both practical and theoretical knowledge. The psychologist Rebecca Lawson (2006, p. 1674) tells a remarkable personal anecdote:

> I regularly scuba dive, and on one weekend trip two friends and I decided to do a night dive. We needed our weight belts, which were on a boat in the harbor, so I offered to fetch them. I swam the few meters to the boat in my dry suit, clipped my own weight belt around my waist, held one weight belt in each hand, slipped overboard, and headed back to the ladder. To my surprise, I immediately sank to the bottom of the harbor. I flailed up and managed to gasp some air but could not move forward before I was dragged down again. Realizing that I was well on my way to drowning, I dumped the two loose weight belts and managed to reach the ladder. In retrospect, the mistake was shocking. I have spent years carrying lead weight belts around: Their heaviness is both perceptually highly salient and central to their function. I had tried to carry three people's weight belts, even though I knew that my own weight belt was adjusted to make me only just buoyant in a dry suit. Why, then, did I fail to anticipate what would happen when I jumped into the water carrying enough lead to sink me, no matter how hard I swam? My conceptual knowledge of weight belts let me understand what had happened after the event, but it was not until the physical constraints of the real world impinged that I accessed this information. Such post hoc explanations created after perceptual experiences allow us to learn from our mistakes, but otherwise may have little influence on our everyday actions.

A theory in which all physical reasoning was done by a veridical simulation that was accurate up to limits posed by perceptual noise and limits on working memory would offer little insight into why in many physical domains errors seem so pervasive.

## 4.2 Intermittent use of a correct physical theory

A challenge to a pure simulation account comes from the fact that people's accuracy can vary widely depending on the framing of a given problem, even if the underlying physics is essentially identical. Indeed, in some cases, two tasks that differ only in the desired form of the answer can yield entirely different results. For example, when subjects are asked to reason about the flight of the bob of a pendulum after it is cut, participants do well if they are asked to place a basket so as to catch the basket or to place a blade so the bob will reach a specified spot (Smith and Vul (2013)), but in the same study it was found that participants do poorly if they are asked to draw the trajectory of the bob.

To take another example, Battaglia, Hamrick, and Tenenbaum (2013) propose that reasoning about falling towers of blocks involves a full Newtonian theory of solid objects. However, in other settings governed by the same physical theory, people do very badly. The principle underlying a balance beam, for instance, involve only a rather simple subset of Newtonian physics; but it is well established that naïve subjects make systematic basic errors in reasoning about balance beams (Siegler, 1976) and we have shown that these errors cannot be accounted for using the model of imperfect perception in (Battaglia, Hamrick, & Tenenbaum, 2013). Note that a balance beam can be built as a kind of tower, and in fact will, with some probability, be built by the "random tower construction" algorithm that Battaglia, Hamrick, and Tenenbaum are using to generate test examples. Similarly, Newtonian theory of solid objects, and the specific physics engine used by Battaglia, Hamrick, and Tenenbaum also incorporates gyroscopic motion; but naïve subjects not cannot accurately predict the behavior of a gyroscope; indeed, they can hardly believe that a gyroscope functions as it does even when they directly experience it.

As a result, the predictive power of the "noisy Newton" theory is severely limited. It is not the case that people reliably use a noisy Newton method, or any kind of Newton-based method, for problems involving rigid solid objects. It is not even the case that they use reliably these methods for the specific problem of predicting the behavior of cutting pendulums. Until there is a characterization of which problems invoke "noisy Newton", the theory does not make any general, testable predictions.

## 4.3 Implausibly rich tacit theories

As Sanborn, Mansinghka and Griffiths (2013) acknowledge, "Newtonian physics and intuitive physics might seem far apart", given that the "The discovery of Newtonian physics was a major intellectual achievement and its principles remain difficult for people to learn explicitly". Of course, we know that people can have tacit knowledge that can be very difficult to make explicit; for instance linguists struggle to characterize language, and children manage to learn language readily without being able to articulate the underlying rules. Nonetheless, both the gap between people's performances on physical reasoning tasks and Newtonian physics, and the inherent difficulty of characterizing the complexities of the real world in terms of scientific principles shed doubt on claims that human physical reasoning largely reflects correct scientific theories.

Consider, for example, the motions of bicycles; although they are familiar to most people, it turns out that the full theory of bicycle motion was not correctly understood until 2011 (Schwab & Meijaard, 2013).[4] It is of course implausible that we are innately endowed with such a theory, since there were no bicycles in the principle period of human evolutionary adaptation, yet it also hardly seems plausible that the average teenager has worked out something that eluded physicists for so long. Instead, it seems more plausible to assume that our notion of bicycle mechanisms revolves around some sort of intuitive simplification that derives from experience.

To take another example, consider the process of cooking scrambled eggs. To what extent the physical chemistry that explains the change in physical characteristics and taste wrought by heat and stirring in turning a raw egg into a scrambled egg is currently understood we do not know, but it seems altogether implausible to suppose that naïve subjects have tacit knowledge of this physical chemistry. A subject who has never seen an egg being cooked would be very unlikely to be able to predict its behavior from first principles. Cooks seems to know "one-off" characteristics of eggs specifically, via experience, not an instance of any useful more general theory.

On the other hand, people are able to connect their very imperfect understanding of cooking eggs with their physical reasoning generally. They know, for example, that they can flick a piece into the air with a fork; and that if they turn the pan upside down, the eggs will fall out, even if they have never tried it. They can guess that if you mix blue ink into them they will probably turn blue; and that they can probably cook an ostrich egg the same way, but it will probably take longer. This ability to combine theories of very different levels of detail is very difficult to attain with physics engines, since a physics engine requires a precise theory; alternative reasoning techniques such as symbolic reasoning are much more flexible in that regard.

## 4.4 Simulation as the result, rather than the mechanism, of physical reasoning

In a number of cognitive studies of visualization or embodied simulation, it is clear on consideration that most of the physical reasoning involved must be carried out *before* the simulation can be constructed; and that therefore, the simulation cannot be the cognitive mechanism that supports the physical reasoning. For example, Zwaan and Taylor (2006) report an experiment in which subjects read one of the following two texts:

    1) The carpenter turned the screw. The boards had been connected too tightly.
    2) The carpenter turned the screw. The boards had been connected too loosely.

---

[4] This despite the fact that a bicycle is a quite simple system; in terms of a stability analysis, a bicycle with two wheels on the ground consists of four rigid objects (the two wheels, the handlebar, and the frame) with six degrees of freedom, and numerous symmetries.

Subjects who read sentence (1) then find it easier to turn their hand in a counterclockwise direction, while subjects who read sentence (2) find it easier to turn their hand clockwise. But the connections between the texts and the directions of turning the screw themselves rest on a rather complex sequence of inference, combining reasoning about the physics with reasoning about the carpenter's purposes

Similar considerations apply to non-physical visualizations as well. For instance, Moulton and Kosslyn (2009) discuss a hypothetical case of a groom who is considering telling a risqué wedding toast, visualizes the look of horror on the bride's face, and changes his mind. But how does the groom come to *visualize* his bride with a look of horror? The real work is beforehand; the cognitive process that generates the visualization must surely be drawing on a previous inference that the bride will be horrified based on knowledge of speech acts, ceremonial occasions, emotions, and so on. It is not plausible that there is a process that goes directly from the joke to the *image* of horror and then interprets the image to infer the emotion of the bride.

## 5. Memory-based physical reasoning

An alternative theory of constructing simulations is that we don't literally simulate, but instead effectively playback movies of prior experiences, retrieving relevant memories and modifying them to fit the current situation. For instance if you see an open thermos of hot chicken soup knocked off a table edge, you recall other cases where you have seen containers of liquid spill and predict that the results here will be similar. Of course, if you do no step-by-step iterative computation, simulation is not involved; you have used a shortcut, not a simulation.

Sanborn, Masingkha, and Griffiths (2013) propose one version of more simulative theory, in which a "noisy Newton" model of collisions could be implemented using stored memories of observed collisions. When a new collision must be reasoned about, the reasoner would retrieves related stored memories and evaluates the new situation by a process of interpolation among the stored memories, taking into account uncertainty in the perception of the new situation.

The theory initially seems plausible, and we do not doubt that in some cases retrieval from memory plays a significant part of the reasoning process.[5]

But in many situations, the machinery of memory per se is too impoverished to yield clear answers, without significant support from other mechanisms. Consider, for example, Schwartz and Black's (1996) gear problem. Since the problem involves this particular pair of gears, with a dent and with a bump, which the subjects have presumably never seen before, they certainly cannot retrieve a memory that gives the answer to this precise question. It even seems unlikely that subjects in general have a memory of seeing the interaction of one gear with a bump and another with a corresponding dent. Subjects presumably have seen gears interact (otherwise, they could hardly answer the question), and could retrieve memories of seeing those. But in order to apply that memory to the current problem, they must be generalize the experience: either to the narrow rule, "Two interacting gears of the same radius will rotate with the same angular velocity" or to the broader rule, "Two interacting gears rotate in such a way that the contact point moves at equal speeds along the two boundaries," or something similar. And since they will need to carry out this generalization in any case, there is not much reason, either experimental or theoretical, to presume that the generalization is performed at query time; it would seem much more effective to carry it out when the subject learns about gears, and then have it available. Memory per se contributes little.

---

[5] The machine learning techniques of "nearest-neighbors" and "case-based reasoning" are both implementations of this general idea.

Second, finding the correct measure of similarity between a new situation and previous memories and finding the proper way to modify an earlier memory to fit a new situation often require substantial background knowledge and reasoning. Consider again the example of knocking over a thermos of soup, and imagine a subject who has often seen a thermos of soup but, as it happens, has never seen either a thermos or soup spill, though he has seen cups of coffee, glasses of wine, glasses of juice, pots of water and so on spill. The subject must succeed in retrieving the experience of seeing other open containers of liquid spill as the relevant memory rather than the experience of seeing thermoses of soup not spill (because they remain upright). Then the subject must determine how the features of the current situation combine with those of the memories; e.g. there will be a puddle of chicken soup rather than a puddle of coffee or wine; if it falls onto fabric, it will create a grease stain rather than a brown or red stain. A glass thermos may break, like a wine glass; a metal thermos will probably not break, like a paper coffee cup. We are not arguing that these inferences cannot be carried out properly; indeed there is a substantial literature on how that might be done; but memory alone is hardly a panacea. At the very least carrying out the inferences require integrating the memory retrieval system with some knowledge-based reasoning; unaided by other mechanisms, memory would be of little use.

The issue of similarity in comparing events is studied in depth in (Lamberts, 2004). Lamberts points out that the proper measure of similarity between two events, viewed as trajectories of objects, is rather obscure; similarity judgments are, to a substantial extent, invariant under changes of both spatial and temporal scale, and dominated by qualitative characteristics such as linear vs. curvilinear motion. The larger problem however, not mentioned by Lambert, is that the relevant measure of similarity depends on the *kind* of prediction being made. In predicting the trajectory of motion, a memory of coffee spilling from a cup is a good exemplar for chicken soup spilling from a cup; in predicting the color of a stain and its response to detergent, a memory of dripping chicken soup from a spoon would be a better exemplar. Echoing points made above, the real work is left to the engine that determines what kind of memories to search for, relative to the type of prediction being made, rather than the machinery of memory per se.

Similar issues arise in using memory-based simulation for predicting of the behaviors towers of blocks, considered in (Battaglia, Hamrick, & Tenenbaum, 2013). On first glance, this might appear a natural application for this technique; one simply remembers a similar tower and predicts a similar behavior, or a number of similar towers and have them vote on the behavior. But it is easily shown that the obvious simple measure of "similarity" between towers of blocks — match up the blocks in a one-to-one correspondence, and then measure the closeness of parameters such as the coordinates of the centers of the blocks and their dimensions — will give poor results unless the subject has seen a vast number of towers[6], and even so the theory does not yield cognitively plausible errors. First, the critical parameters here are not the coordinates or dimensions directly; they are the position of the centers of mass of each subtower relative to the boundaries of the supporting block. Second, the remembered tower that is actually most relevant to the current tower may not match up block by block; a single block in one may correspond to multiple blocks in the other. (Figure 7).

---

[6] With a large enough data set , nearest neighbors can be surprisingly effective, even for tasks as complex as image categorization (Torralba, Fergus, & Freeman, 2008).

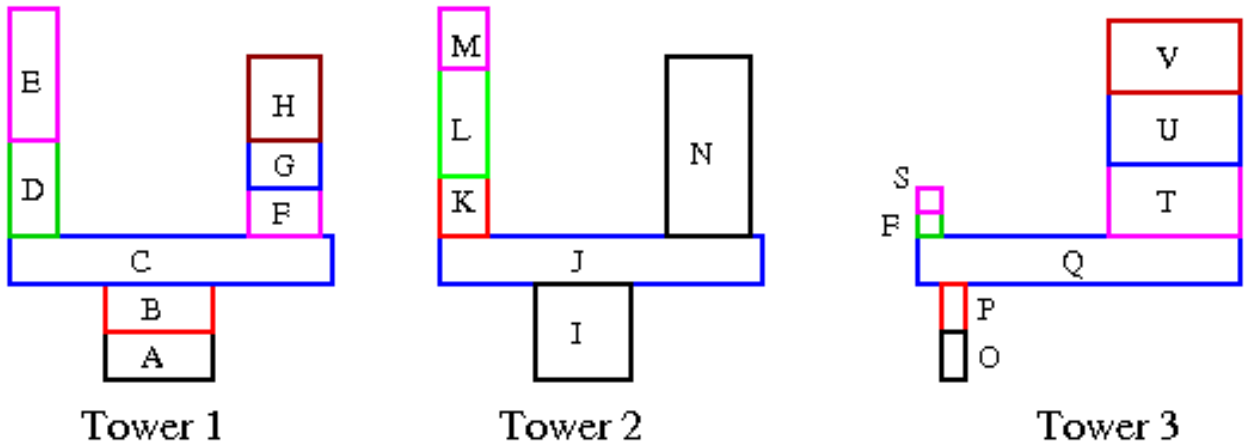

**Figure 7: Tower 1 is a better exemplar for Tower 2 than for Tower 3, despite the fact that there is a one-one matching between the blocks in Towers 1 and 3, preserving support and left-right relations, whereas the matching between Tower 1 and Tower 2 is many-many (A,B match I; D,E match K,L,M; F,G,H match N).**

Another difficult problem is the need to merge multiple separate memories. Consider, for example, a subject who sees an open cup of coffee perched on top of an unstable tower of blocks. To predict what will happen, it is necessary to combine previous memories of falling towers of blocks with memories of spilling containers of liquids. In a knowledge-based system, such combinations are generally fairly straightforward, though certainly difficulties can arise; in a physics engine-based system, this can easy, if the engine already incorporates all interactions, or laborious, if it does not; but in a memory-retrieval system it is generally highly problematic.

The third issue is that many (though not all) of the problems we have described in physics engines turn up, in altered forms, in retrieval of remembered experiences as well. The problem of finding an appropriate model is substantially reduced, since the reasoner often does not have to understand a behavior, merely to have witnessed or experienced it. However, all the other problems arise, or have very close analogues. The problem of finding the correct idealization is essentially the same as the problems, discussed above, of finding the correct measure of similarity and the correct way to modify a memory. The problems of incomplete information is a well-known stumbling block in nearest-neighbor algorithms and, to a lesser extent, in case-based reasoning; if the value of a feature is not known, it is hard to know how to include it in the measure of similarity. The problems of easy inference, of incorporating non-physical information and of non-predictive tasks are also all essentially unchanged.

In the final analysis, memory may well play an important role in physical reasoning, but it cannot solve the problems on its own; even if an organism had a perfect record of all its experiences, immediately accessible at any moment, there would still be major challenges in determining which memories were relevant at any given moment, and how to generalize and adapt them to new situations.

# 6. Non-veridical simulation

## 6.1 Partially specified models, supplemented with machinery borrowed from formal logic.

A critic might respond to some of the points we have raised above by claiming that simulations do not have to be veridical; they can be cartoons or schemas that stand in for the world, rather than exact simulations of physical reality. SimCity doesn't need to have a one-to-one correspondence to real world financial institutions in order to be an orderly, internally coherent proxy that stands in for the real world and teaches children something about capitalism. Johnson-Laird (1983) even argues that mental models do not have to be fully specified; they can include unspecified elements. If our critique applied only to fully-specified veridical simulations, it would not prove very much.

Once one start to introduce partially specified models, however, the distinction between models and alternatives such as inference using propositional logic starts to become hazy; the more flexible the kinds of partial specification allowed in the models, the hazier the distinction becomes. Johnson-Laird, for example, allows a symbol for "not". If one additionally allows symbols for "and" and "exists" and a scoping mechanism, the system becomes equivalent to first-order logic,. At that point, it becomes unclear what distinguishes a "simulation"-based model from any other inference technique. The computational advantages of simulation and virtually all the empirical evidence in favor of simulation in cognition are grounded in an interpretation of simulation as an operation on fully specified descriptions of the world. If that is weakened, the theory becomes much more nebulous.

## 6.2 Non-realistic diagrams.

People often find diagrams extremely useful in carrying out physical reasoning tasks. Physical situations that cannot be understood from any verbal description can be immediately apprehended given a picture, or, better yet, a video. The natural explanation of this is that useful geometric information that would be difficult to compute from a symbolic representation can be directly computed from a physical picture or from a picture-like internal representation. For example, Chadrasekaran, Glasgow, and Narayanan (1995, p. xxii) write, "Diagrams preserve, or represent directly, locality information. A number of visual predicates are efficiently computed by the visual architecture from the above information, e.g. neighborhood relations, relative size, intersections, and so on. This ability makes certain types of inferences easy or relatively direct."

This is a plausible explanation of realistic drawings. An engineer or architect, for example, finds it useful to construct a scale drawing or model because all geometric relations can simply be measured, and any error that has been made in geometric calculation is immediately apparent. The problem with the explanation, however, is that people find non-veridical diagrams just as useful, in the same way, and in those cases the explanation is much more problematic. Consider, for example, figure 7, from the Wikipedia article, "Redshift", which shows a star, a light beam, and an observer. This diagram is useful, but not veridical, given the extreme distortion of relative scale.

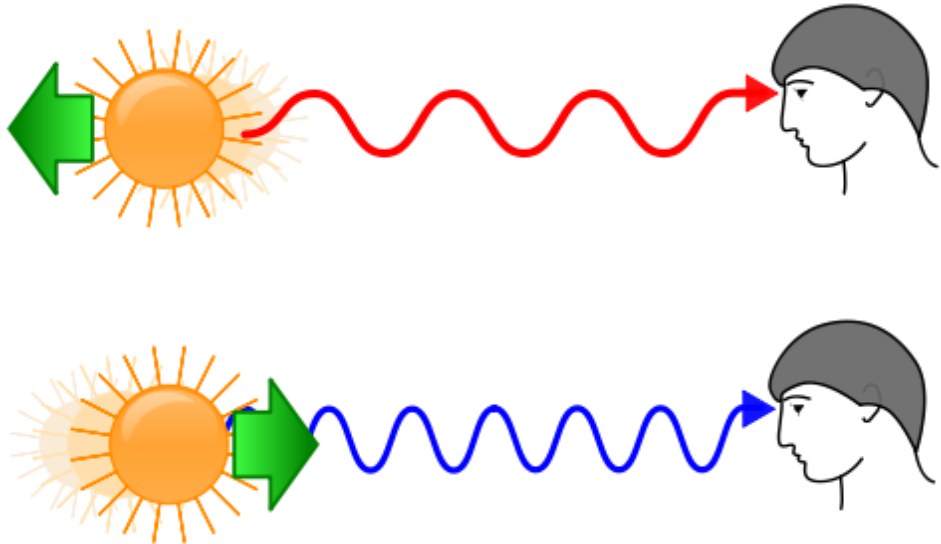

"Redshift and blueshift", by Aleš Tošovský from "Redshift", Wikipedia

**Figure 8: Star, Light, Observer**

The distance from the star to the observer is shown as equal to 4 wavelengths of red light; in reality, it is $4*10^{24}$ wavelengths. Note that the wavelength is not purely an icon, like the big green arrows next to the star; the comparative wavelength of the blue and the red lights is meaningful and important to the point being illustrated. The scale of the star and the scale of the observer are likewise hugely distorted.

Similarly, figure 9, from Feynman's *Lectures on Physics*, illustrates molecules of gas in a piston.

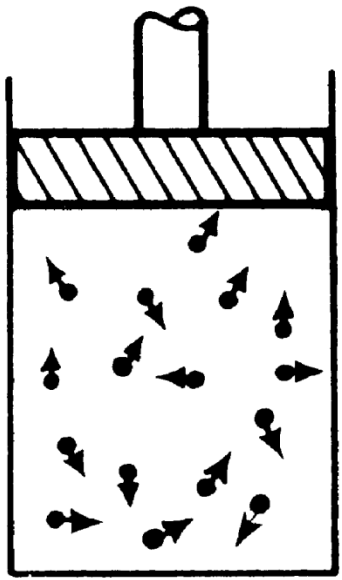

**Figure 9: Gas molecules in a piston**

In the actual situation, there are of course about $10^{23}$ molecules of gas.

The trouble is that if diagrams are taken literally, serious problems start to arise in distinguishing conclusions that are true in the real situation from those that are merely artifacts of the diagram. For instance, if one took figure 8 literally, one would conclude incorrectly that, since the light is shown as goes up and down on a path with a 4 inch amplitude, a 3x5 index card will sometimes fail to block one's view of an object directly ahead (even when the card is held directly in front of the eyes), and that the card would sometimes succeed in blocking it, even when the card is held 4 inches below the line of sight (figure 10). There is nothing in figure 8 to rule this out; it must be ruled out by reasoning that lies outside this picture. Similarly the student looking at figure 9 is supposed to find it helpful in understanding that the motion of the molecules exert pressure on the inside of the piston, but they are not supposed to draw the invalid conclusion that the gas molecules collide much more often with the sides of the piston than with one another, which would be true in the situation depicted. Non-veridical simulations raise difficult problems of interpretation that can often only be solved using non-simulative reasoning. And if diagrams are not taken literally, the role of simulation in a larger theory once again becomes greatly diminished.

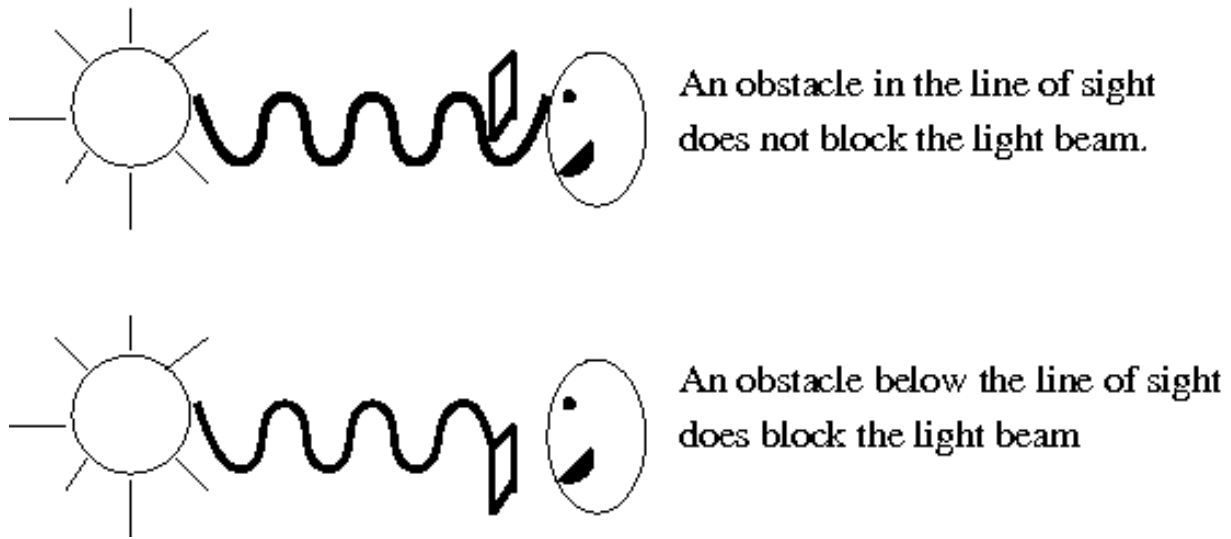

**Figure 10: Invalid inferences from figure 8**

# 7. Conclusions

Simulation-based theories of physical reasoning are both too weak and too strong. They are too weak in that there are many forms of physical reasoning that people carry out where simulation is either extremely inefficient or entirely inapplicable. They are too strong, at least in the "noisy Newton" formulation, because in many tasks they predict much higher levels of performance than people exhibit; only in rare cases can the large, qualitative, fundamental, systematic errors that humans make be explained in terms of perceptual noise.

At this juncture, it is difficult or impossible to quantify what fraction of the physical reasoning in human cognition or in a general artificial intelligence is or could be carried out using simulation. We have argued, however, that the range of examples we have presented in this paper — many constructed as simple variants of problems considered in the pro-simulation literature — suggests that there are significant limits to the use of simulation. In particular, although we have suggested that simulation is effective for physical reasoning when the task is prediction, when complete information is available, when a reasonably high quality theory is available, and when the range of spatial or temporal scale involved is moderate, in many other cases simulation is to some extent problematic. In particular, physical reasoning often involves tasks other than prediction and information is often partial.

Moreover, even when optimal conditions hold, there are many cases in which it would appear that alternative non-simulative modes of reasoning are likely to be easier, faster, or more robust. Finally, setting up and interpreting a simulation requires modes of physical reasoning that are not themselves simulation. For all these reasons, we suggest that non-simulative forms of reasoning are not an optional extra in cognitive theories but are centrally important.

One reason that researchers have overstated the role of simulation in cognitive models is that in the majority of studies of physical reasoning in the psychological literature, subjects are asked to carry out task of prediction and in the overwhelming majority they are presented with complete specifications of the situation.[7] However, there is little reason to suppose that that at all reflects the frequency of these forms of reasoning in ecologically realistic settings; it may well result from artificial constraints that arise in setting up a controlled experiment.

We concur with Hegarty (2004) in seeing room for hybrid models that combine simulation with other techniques, such as knowledge-based inference,. In our view, however, simulation may play a much less central role than Hegarty envisioned In particular, in a general intelligence, human or artificial, practically any reasoning that involves simulation will probably involve some degree of non-simulative reasoning, to set up the simulation and to check that the answer is broadly reasonable. Outside of carefully constructed laboratory situations, and certain forms of specialized expert reasoning, we think it is relatively rare that simulation is used in isolation in human cognitive reasoning.

If there is indeed a physics engine in the head, it is likely to be only a small part of a larger system that encompasses a wide range of additional cognitive processes, such as learning, memory-based reasoning, causal reasoning, qualitative reasoning, rule-based heuristics, analogy, and abstraction, ,

## Acknowledgements

Thanks to Peter Battaglia, Benjamin Bergen, Thomas Bräunl, Alan Bundy, Jonathan Goodman, Philip Johnson-Laird, Owen Lewis, Casey McGinley, Andrew Sundstrom, and Ed Vul for helpful discussions; and to Nils Thuerey for information about simulation for film CGI.

---

[7] There is also a tendency in the cognitive science literature to focus on idealized physical models that are characteristic of textbooks or engineered devices, such as balance beams, perfectly bouncing balls, and gears rather than those that occur in natural settings like camping trips or kitchens.